\pdfoutput=1

\documentclass[11pt]{article}

\usepackage[table,xcdraw]{xcolor}

\usepackage[]{EMNLP2022}

\usepackage{times}
\usepackage{latexsym}

\usepackage[T1]{fontenc}

\usepackage[utf8]{inputenc}

\usepackage{microtype}

\usepackage{inconsolata}

\usepackage{microtype}
\usepackage{graphicx}
\usepackage{amsmath} 
\usepackage{amsfonts}
\usepackage{booktabs}
\usepackage{multirow}
\usepackage{multicol}
\usepackage{subfig}
\usepackage[ruled,linesnumbered,boxed]{algorithm2e}
\usepackage{hyperref}
\hypersetup{
hidelinks,
colorlinks=true,
allcolors=black,
pdfstartview=Fit,
breaklinks=true
}

\title{Distilling Causal Effect from Miscellaneous Other-Class\\ for Continual Named Entity Recognition}

\author{Junhao Zheng, Zhanxian Liang, Haibin Chen, Qianli Ma*\\
  School of Computer Science and Engineering, \\
  South China University of Technology, Guangzhou, China\\
  \texttt{junhaozheng47@outlook.com}, 
  \texttt{qianlima@scut.edu.cn}\thanks{*Corresponding author}}
  
\begin{document}
\maketitle
\begin{abstract}
Continual Learning for Named Entity Recognition (CL-NER) aims to learn a growing number of entity types over time from a stream of data.
However, simply learning $\textit{Other}$-Class in the same way as new entity types amplifies the catastrophic forgetting and leads to a substantial performance drop.
The main cause behind this is that $\textit{Other}$-Class samples usually contain old entity types, and the old knowledge in these $\textit{Other}$-Class samples is not preserved properly.
Thanks to the causal inference, we identify that the forgetting is caused by the missing causal effect from the old data.
To this end, we propose a unified causal framework to retrieve the causality from both new entity types and $\textit{Other}$-Class.
Furthermore, we apply curriculum learning to mitigate the impact of label noise and introduce a self-adaptive weight for balancing the causal effects between new entity types and $\textit{Other}$-Class.
Experimental results on three benchmark datasets show that our method outperforms the state-of-the-art method by a large margin.
Moreover, our method can be combined with the existing state-of-the-art methods to improve the performance in CL-NER. \footnote{Our codes are publicly available at \href{https://github.com/zzz47zzz/CFNER}{https://github.com/zzz47zzz/CFNER}}
\end{abstract}

\section{Introduction}
Named Entity Recognition (NER) is a vital task in various NLP applications \citep{ma2016end}.
Traditional NER aims at extracting entities from unstructured text and classifying them into a fixed set of entity types (\textit{e.g.}, \textit{Person}, \textit{Location}, \textit{Organization}, \textit{etc}).
However, in many real-world scenarios, the training data are streamed, and the NER systems are required to recognize new entity types to support new functionalities, which can be formulated into the paradigm of continual learning  (CL, \textit{a.k.a.} \textit{incremental learning or lifelong learning}) \citep{thrun1998lifelong,parisi2019continual}.
For instance, voice assistants such as Siri or Alexa are often required to extract new entity types (\textit{e.g.} \textit{Song}, \textit{Band}) for grasping new intents (\textit{e.g.} \textit{GetMusic}) \citep{monaikul2021continual}.

However, as is well known, continual learning faces a serious challenge called \textit{catastrophic forgetting} in learning new knowledge \citep{mccloskey1989catastrophic,robins1995catastrophic,goodfellow2013empirical,kirkpatrick2017overcoming}.
More specifically, simply fine-tuning a NER system on new data usually leads to a substantial performance drop on previous data. 
In contrast, a child can naturally learn new concepts (\textit{e.g.,} \textit{Song} and \textit{Band}) without forgetting the learned concepts (\textit{e.g.,} \textit{Person} and \textit{Location}).
Therefore, continual learning for NER (CL-NER) is a ubiquitous issue and a big challenge in achieving human-level intelligence.

\begin{figure}[t]
    \centering
    \includegraphics[width=0.6\linewidth]{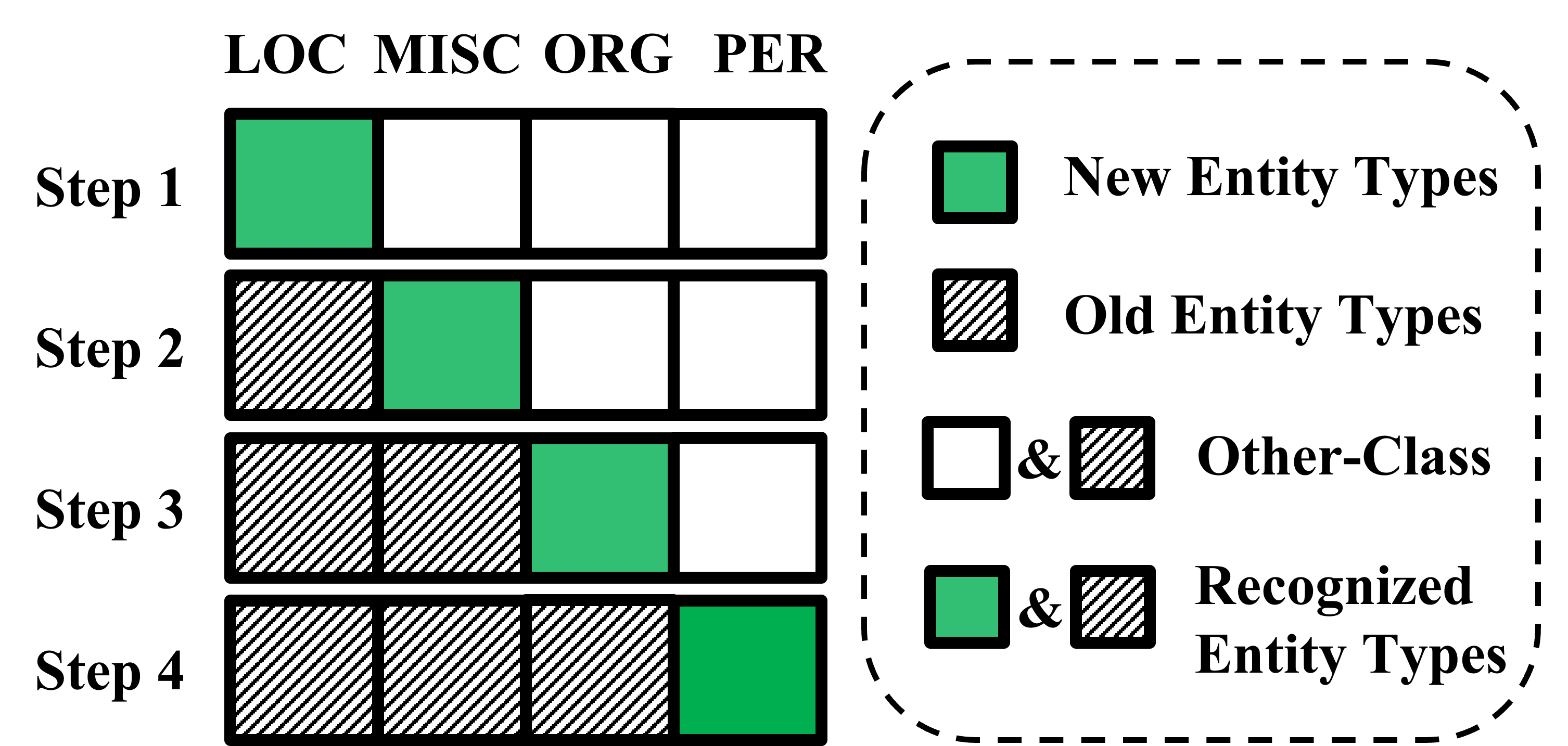}
    \caption{An illustration of \textit{Other}-class in CL-NER. Suppose that a model learns four entity types in CoNLL2003 sequentially. ``LOC'': Location; ``MISC'': Miscellaneous; ``ORG'': Organisation; ``PER'': Person.}
    \label{fig6}
\end{figure}

In the standard setting of continual learning, only new entity types are recognized by the model in each CL step. 
For CL-NER, the new dataset contains not only new entity types but also \textit{Other}-class tokens which do not belong to any new entity types.
For instance, about 89\% tokens belongs to \textit{Other}-class in OntoNotes5 \citep{hovy2006ontonotes}.
Unlike accuracy-oriented tasks such as the image/text classification, NER inevitably introduces a vast number of \textit{Other}-class samples in training data.
As a result, the model strongly biases towards \textit{Other}-class \citep{li2020dice}.
Even worse, the meaning of \textit{Other}-class varies along with the continual learning process.
For example, ``Europe'' is tagged as \textit{Location} if and only if the entity type \textit{Location} is learned in the current CL step.
Otherwise, the token ``Europe'' will be tagged as \textit{Other}-class.
An illustration is given in Figure \ref{fig6} to demonstrate \textit{Other}-class in CL-NER.
In a nutshell, the continually changing meaning of \textit{Other}-class as well as the imbalance between the entity and \textit{Other}-class tokens amplify the forgetting problem in CL-NER.

\begin{figure}[t]
    \centering
    \includegraphics[width=0.6\linewidth]{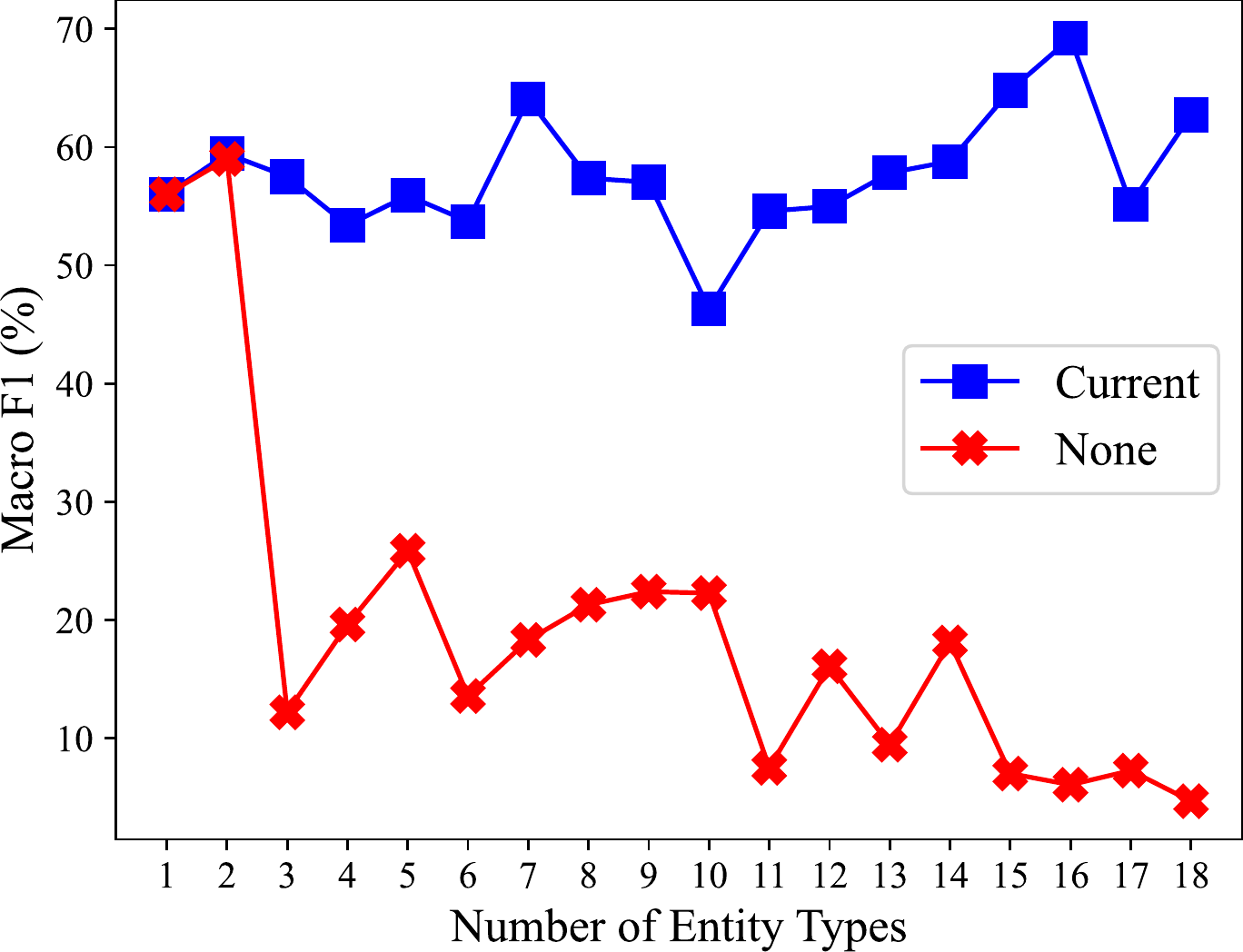}
    \caption{An illustration of the impact of \textit{Other}-class samples on OntoNotes5. We consider two scenarios with different extra annotation levels on \textit{Other}-class samples: (1) annotate all recognized entity types on the data in the current CL step (Current); (2) no extra annotations on \textit{Other}-class samples (None).}
    \label{fig1}
\end{figure}

Figure \ref{fig1} is an illustration of the impact of \textit{Other}-class samples.
We divide the training set into 18 disjoint splits, and each split corresponds to one entity type to learn.
Then, we only retain the labels of the corresponding entity type in each split while the other tokens are tagged as \textit{Other}-class.
Next, the NER model learns 18 entity types one after another, as in CL.
To eliminate the impact of forgetting, we assume that all recognized training data can be stored.
Figure \ref{fig1} shows two scenarios where \textit{Other}-class samples are additionally annotated with ground-truth labels or not.
Results show that ignoring the different meanings of \textit{Other}-classes affects the performance dramatically. 
The main cause is that \textit{Other}-class contains old entities.
From another perspective, the old entities in \textit{Other}-class are similar to the reserved samples of old classes in the data replay strategy \citep{rebuffi2017icarl}.
Therefore, we raise a question: \textit{how can we learn from  \textit{Other}-class samples for anti-forgetting in CL-NER?}

In this study, we address this question with a \textbf{C}ausal \textbf{F}ramework for CL-\textbf{NER} (\textbf{CFNER}) based on causal inference \citep{glymour2016causal,scholkopf2022causality}. 
Through causal lenses, we determine that the crux of CL-NER lies in establishing causal links from the old data to new entity types and \textit{Other}-class.
To achieve this, we utilize the old model (\textit{i.e.,} the NER model trained on old entity types) to recognize old entities in \textit{Other}-class samples and distillate causal effects \citep{glymour2016causal} from both new entity types and \textit{Other}-class simultaneously.
In this way, the causality of \textit{Other}-class can be learned to preserve old knowledge, while the different meanings of \textit{Other}-classes can be captured dynamically.
In addition, we design a curriculum learning \citep{bengio2009curriculum} strategy to enhance the causal effect from \textit{Other}-class by mitigating the label noise generated by the old model.
Moreover, we introduce a self-adaptive weight to dynamically balance the causal effects from \textit{Other}-class and new entity types.
Extensive experiments on three benchmark NER datasets, \textit{i.e.}, OntoNotes5, i2b2 \citep{murphy2010serving} and CoNLL2003 \citep{sang2003introduction}, validate the effectiveness of the proposed method.
The experimental results show that our method outperforms the previous state-of-the-art method in CL-NER significantly. The main contributions are summarized as follows:
\begin{itemize}
    \item We frame CL-NER into a causal graph \citep{pearl2009causality} and propose a unified causal framework to retrieve the causalities from both \textit{Other}-class and new entity types.
    \item We are the first to distillate causal effects from \textit{Other}-class for anti-forgetting in CL, and we propose a curriculum learning strategy and a self-adaptive weight to enhance the causal effect in $\textit{Other}$-class.
    \item Through extensive experiments, we show that our method achieves the state-of-the-art performance in CL-NER and can be implemented as a plug-and-play module to further improve the performances of other CL methods.
\end{itemize}

\section{Related Work}
\subsection{Continual Learning for NER}
Despite the fast development of CL in computer vision, most of these methods \citep{douillard2020podnet,rebuffi2017icarl,hou2019learning} are devised for accuracy-oriented tasks such as image classification and fail to preserve the old knowledge in \textit{Other}-class samples.
In our experiment, we find that simply applying these methods to CL-NER does not lead to satisfactory performances.

In CL-NER, a straightforward solution for learning old knowledge from \textit{Other}-class samples is \textit{self-training} \citep{rosenberg2005semi,de2019continual}.
In each CL step, the old model is used to annotate the \textit{Other}-class samples in the new dataset.
Next, a new NER model is trained to recognize both old and new entity types in the dataset.
The main disadvantage of \textit{self-training} is that the errors caused by wrong predictions of the old model are propagated to the new model \citep{monaikul2021continual}.
\citet{monaikul2021continual} proposed a method based on knowledge distillation \citep{hinton2015distilling} called \textit{ExtendNER} where the old model acts as a teacher and the new model acts as a student.
Compared with \textit{self-training}, this distillation-based method takes the uncertainty of the old model’s predictions into consideration and reaches the state-of-the-art performance in CL-NER. 

Recently, \citet{das-etal-2022-container} alleviates the problem of Other-tokens in few-shot NER by contrastive learning and pretraining techniques.
Unlike them, our method explicitly alleviates the problem brought by \textit{Other}-Class tokens through a causal framework in CL-NER. 

\begin{figure*}[!t]
    \centering
    \subfloat[forgetting]{
        \includegraphics[width=0.5\linewidth]{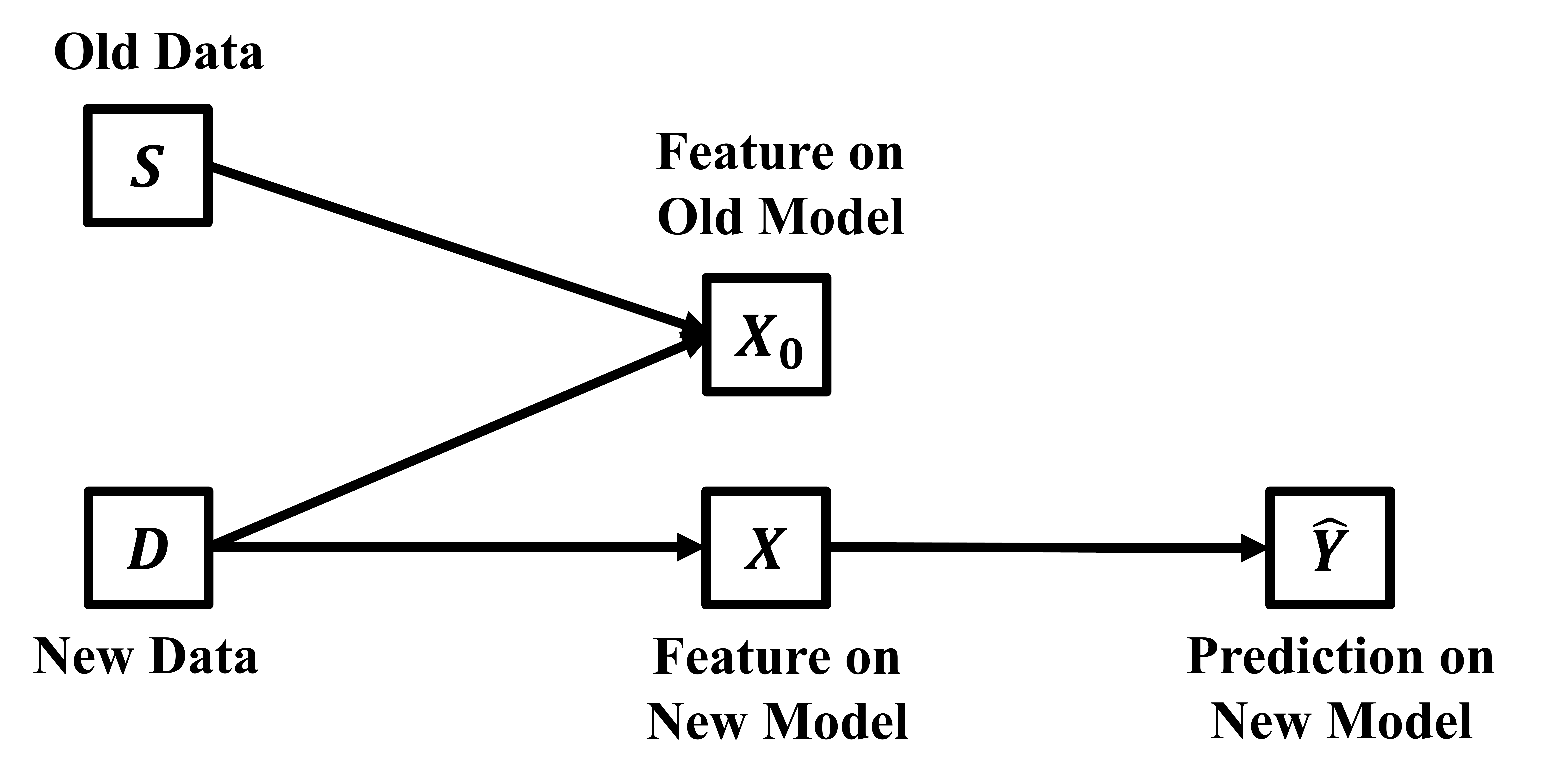}
        \label{fig2a}
    }
    \subfloat[anti-forgetting]{
        \includegraphics[width=0.5\linewidth]{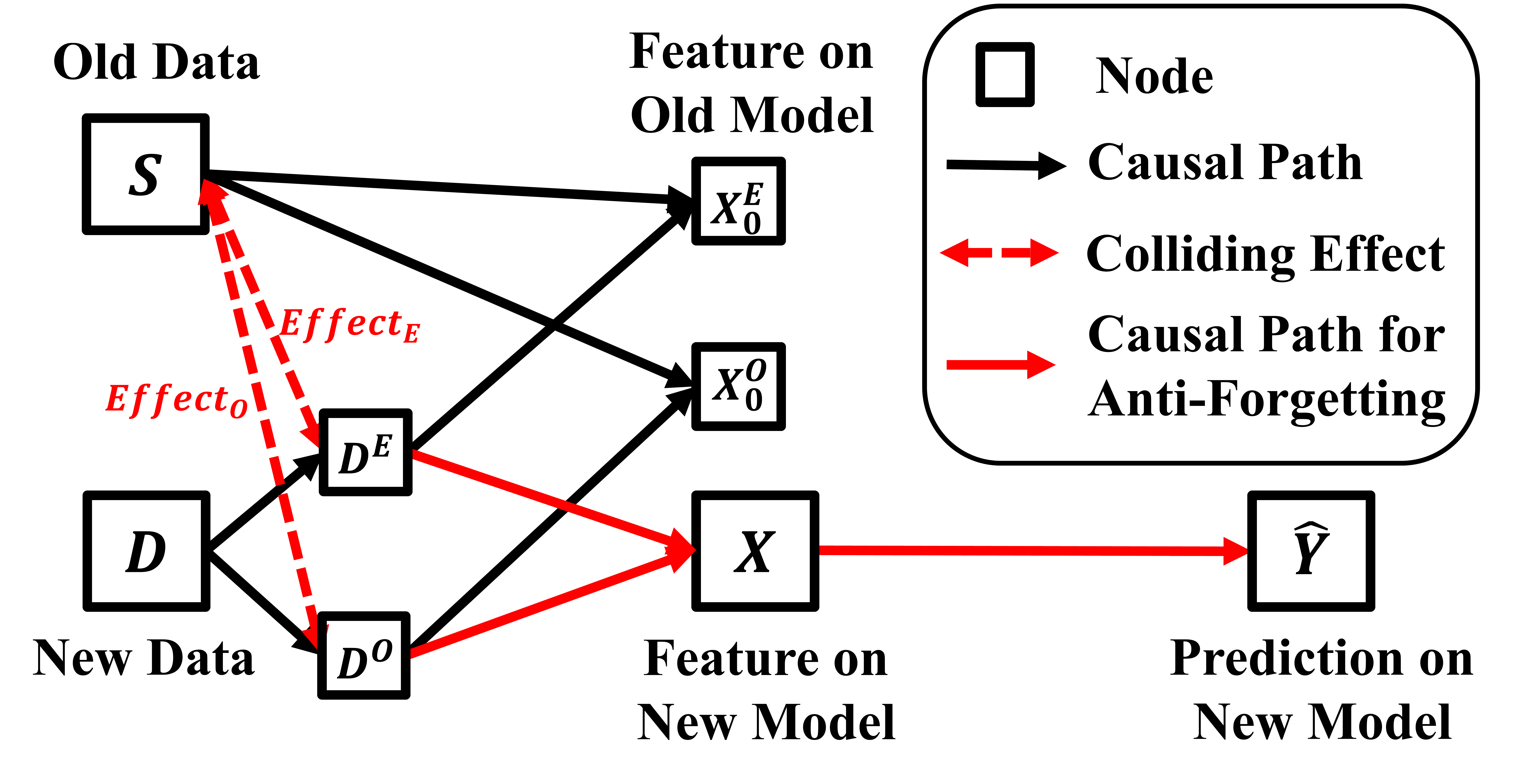}
        \label{fig2b}
    }
    \caption{The causal graph for CL-NER: (a) forgetting happens when there are no causal paths from old data to new predictions; (b) anti-forgetting is to build causal paths from old data to new predictions through new entities ($D^E$) and \textit{Other}-class samples ($D^O$). We call the causal effects in these two links $\textit{Effect}_E$ and $\textit{Effect}_O$, respectively.}
\end{figure*}

\subsection{Causal Inference}
Causal inference \citep{glymour2016causal,scholkopf2022causality} has been recently introduced to various computer vision and NLP tasks, such as semantic segmentation \citep{zhang2020causal}, long-tailed classification \citep{tang2020long,nan2021uncovering}, distantly supervised NER \citep{zhang2021biasing} and neural dialogue generation \citep{zhu2020counterfactual}.
\citet{hu2021distilling} first applied causal inference in CL and pointed out that the vanishing old data effect leads to forgetting.
Inspired by the causal view in \citep{hu2021distilling}, we mitigate the forgetting problem in CL-NER by mining the old knowledge in \textit{Other}-class samples.

\section{Causal Views on (Anti-) Forgetting}
In this section, we explain the (anti-) forgetting in CL from a causal perspective.
First, we model the causalities among data, feature, and prediction at any consecutive CL step with a causal graph \citep{pearl2009causality} to identify the forgetting problem.
The causal graph is a directed acyclic graph whose nodes are variables, and directed edges are causalities between nodes.
Next, we introduce how causal effects are utilized for anti-forgetting.

\subsection{Causal Graph}
Figure \ref{fig2a} shows the causal graph of CL-NER when no anti-forgetting techniques are used.
Specifically, we denote the old data as $S$; the new data as $D$; the feature of new data extracted from the old and new model as $X_0$ and $X$; the prediction of new data as $\hat{Y}$ (\textit{i.e.,} the probability distribution (scores)).
The causality between notes is as follows:
(1) $D \rightarrow X \rightarrow \hat{Y}$: $D \rightarrow X$ represents that the feature $X$ is extracted by the backbone model ($e.g.,$ BERT \citep{devlin-etal-2019-bert}), and $X \rightarrow \hat{Y}$ indicates that the prediction $\hat{Y}$ is obtained by using the feature $X$ with the classifier ($e.g.,$ a fully-connected layer);
(2) $S \rightarrow X_0 \leftarrow D$: these links represent that the old feature representation of new data $X_0$ is determined by the new data $D$ and the old model trained on old data $S$.
Figure \ref{fig2a} shows that the forgetting happens because there are no causal links between $S$ and $\hat{Y}$.
More explanations about the forgetting in CL-NER are demonstrated in Appendix \ref{appendix:forgetting}.

\subsection{Colliding Effects}
In order to build cause paths from $S$ to $\hat{Y}$, a naive solution is to store (a fraction of) old data, resulting in a causal link $S \rightarrow D$ is built.
However, storing old data contradicts the scenario of CL to some extent.
To deal with this dilemma, \citet{hu2021distilling} proposed to add a causal path $S \leftrightarrow D$  between old and new data by using \textit{Colliding Effect} \citep{glymour2016causal}. 
Consequently, $S$ and $D$ will be correlated to each other when we control the collider $X_0$.
Here is an intuitive example: a causal graph $sprinkler \rightarrow pavement \leftarrow weather$ represents the pavement's condition (wet/dry) is determined by both the weather (rainy/sunny) and the sprinkler (on/off).
Typically, the weather and the sprinkler are independent of each other.
However, if we observe that the pavement is wet and know that the sprinkler is off, we can infer that the weather is likely to be rainy, and vice versa.

\begin{figure*}[t]
    \centering
    \includegraphics[width=15cm,height=5cm]{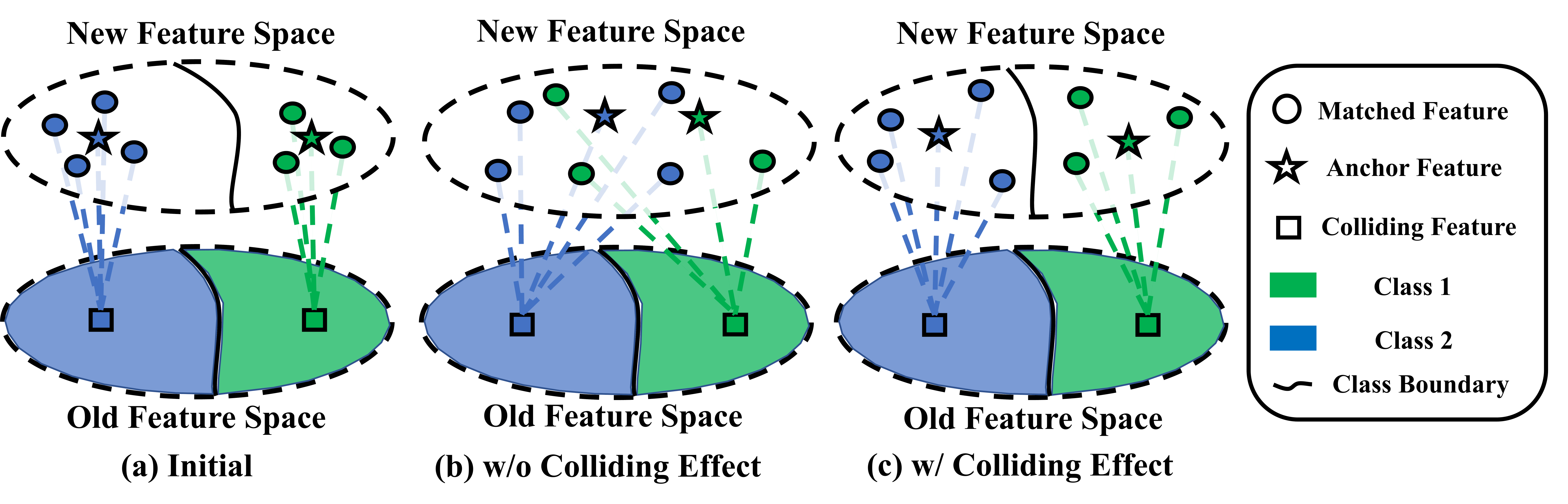}
    \caption{A demonstration of the colliding effect. The anchor token's feature (\textit{anchor features}) collides with matched tokens' features (\textit{matched features}) on the \textit{colliding feature} in the old feature space. (a) initial: the class boundary is retained since the new model is initialized by the old model condition. (b) w/o Colliding Effect: the class boundary is forgot in the new feature space since there are no causal effects from old data to new predictions. (c) w/ Colliding Effect: the class boundary is preserved after an CL step since the anchor and matched tokens \textit{collide} in the old feature space.}
    \label{fig4}
\end{figure*}

\section{A Causal Framework for CL-NER}
In this section, we frame CL-NER into a causal graph and identify that learning the causality in \textit{Other}-class is crucial for CL-NER.
Based on the characteristic of CL-NER, we propose a unified causal framework to retrieve the causalities from both \textit{Other}-class and new entity types.
We are the first to distillate causal effects from \textit{Other}-class for anti-forgetting in CL.
Furthermore, we introduce a curriculum-learning-based strategy and a self-adaptive weight to allow the model to better learn the causalities from \textit{Other}-class samples.

\subsection{Problem Formulation}
In the $i$-th CL step, given an NER model $M_i$ which is trained on a set of entities $E_i=\{ e_1, e_2 \cdots, e_{n_i} \}$, the target of CL-NER is to learn the best NER model $M_{i+1}$ to identify the extended entity set $E_{i+1}=\{ e_1, e_2 \cdots, e_{n_{i+1}} \}$  with a training set annotated only with $E_i^{New} = \{ e_{n_i+1}, e_{n_i+2}, \cdots, e_{n_{i+1}} \}$.

Suppose the model consists of a backbone network for feature extraction and a classifier for classification.
As a common practice, $M_{i+1}$ is first initialized by the parameter of $M_i$, and then the dimensions of the classifier are extended to adapt to new entity types.    
Then, $M_i$ will guide the learning process of $M_{i+1}$ through knowledge distillation \citep{monaikul2021continual} or regularization terms \citep{douillard2020podnet} to preserve old knowledge.
Our method is based on knowledge distillation where the old model $M_i$ acts as a teacher and the new model $M_{i+1}$ acts as a student.
Our method further distillates causal effects in the process of knowledge distillation.

\begin{figure*}[!t]
    \centering
    \includegraphics[width=16cm]{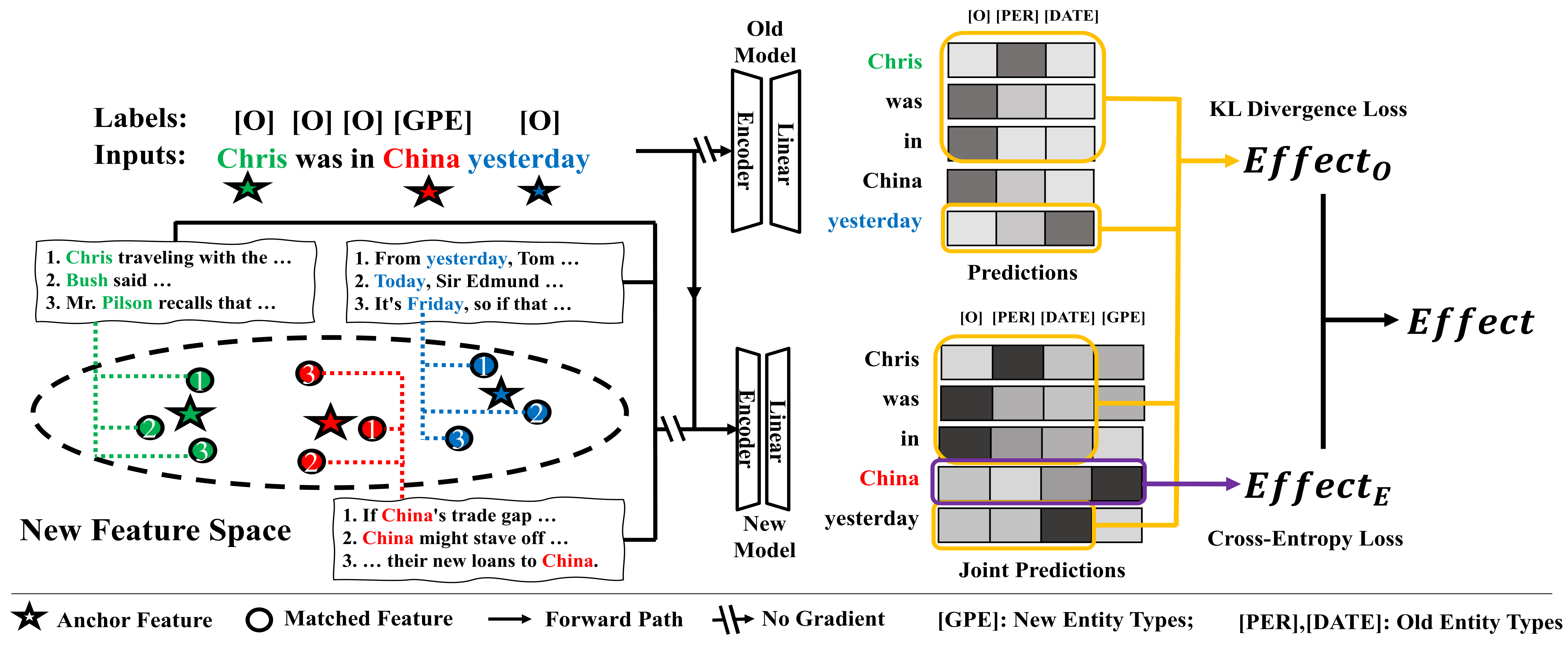}
    \caption{A demonstration of the proposed causal framework for CL-NER.}
    \label{fig5}
\end{figure*}

\subsection{Distilling Colliding Effects in CL-NER}
Based on the causal graph in Figure \ref{fig2a}, we figure out that the crux of CL lies in building causal paths between old data and prediction on the new model.
If we utilize colliding effects, the causal path between old and new data can be built without storing old data.

To distillate the colliding effects, we first need to find tokens in new data which have the same feature representation $X_0$ in the old feature space, \textit{i.e.,} condition on $X_0$.
However, it is almost impossible to find such \textit{matched tokens} since features are sparse in high dimensional space \citep{altman2018curse}.
Following \citet{hu2021distilling}, we approximate the colliding effect using K-Nearest Neighbor (KNN) strategy.
Specifically, we select a token as \textit{anchor token} and search the k-nearest neighbor tokens whose features bear a resemblance to the \textit{anchor token's} feature in the old feature space.
Next, when calculating the prediction of the anchor token, we use \textit{matched tokens} for joint prediction.
Note that in backpropagation, only the gradient of the anchor token is computed.
Figure \ref{fig4} shows a demonstration for distilling colliding effects.

Although \textit{Other}-class tokens usually do not directly guide the model to recognize new entity types, they contain tokens from old entity types, which allow models to recall what they have learned.
Naturally, we use the old model to recognize the \textit{Other}-class tokens which actually belong to old entity types.
Since these \textit{Other}-class tokens belong to the predefined entity types, we call them as \textit{Defined-Other}-Class tokens, and we call the rest tokens in \textit{Other}-class as \textit{Undefined-Other}-Class tokens.

Based on the characteristics of NER, we extend the causal graph in Figure \ref{fig2a} to Figure \ref{fig2b}.
The key adjustment is that the node of new data is split into two nodes, including new entity tokens $D^E$ and \textit{Defined-Other}-Class tokens $D^O$.
Then, we apply the colliding effects on $D^E$ and $D^O$ respectively, resulting in that $D^E$ and $D^O$ \textit{collide} with $S$ on nodes $X_0^E$ and $X_0^O$ in the old feature space.
In this way, we build two causal paths from old data $S$ to new predictions $\hat{Y}$.
In the causal graph, we ignore the token from \textit{Undefined-Other}-Class since they do not help models learn new entity types or review old knowledge.
Moreover, we expect the model to update instead of preserving the knowledge about \textit{Other}-class in each CL step. 
Here, we consider two paths separately because the colliding effects are distilled from different kinds of data and calculated in different ways.

\subsection{A causal framework for CL-NER}
Formally, we define the total causal effects $\textit{Effect}$ as follow:
\begin{align}
 \textit{Effect} &= \textit{Effect}_{E} + \textit{Effect}_{O} \label{eq:total_objective_1} \\
  &= - \Sigma_i \textit{CE}(\overline{Y}_i, Y_i) - \Sigma_j \textit{KL}(\overline{Y}_j,\widetilde{Y}_j), \label{eq:total_objective_2}\\
  \nonumber&\textit{s.t.}\quad D_i \in D^E,\quad D_j \in D^O. 
\end{align}
In Eq.(\ref{eq:total_objective_1}), $\textit{Effect}_{E}$ and $\textit{Effect}_{O}$ denote the colliding effect of new entity types and \textit{Defined-Other}-Class. 
In Eq.(\ref{eq:total_objective_2}), $\textit{CE}(\cdot,\cdot)$ and $\textit{KL}(\cdot,\cdot)$ represent the cross-entropy and KL divergence loss, and $D_i$,$D_j$ are the $i$-th,$j$-th token in new data.
In cross-entropy loss, $Y_i$ is the ground-truth entity type of the $i$-th token in new data.
In KL divergence loss, $\widetilde{Y}_j$ is the soft label of the $j$-th token given by the old model over old entity types.
In both losses, $\overline{Y}$ represents the weighted average of prediction scores over \textit{anchor} and \textit{matched} tokens.
When calculating $\textit{KL}(\cdot,\cdot)$, we follow the common practice in knowledge distillation to introduce temperature parameters $T_t$, $T_s$  for teacher (old model) and student model (new model) respectively.
Here, we omit $T_t$, $T_s$ in $\textit{KL}(\cdot,\cdot)$ for notation simplicity.
The weighted average scores of the $i$-th token is calculated as follow:
\begin{align}
    \overline{Y}_i &= W_i \hat{Y}_i + \Sigma_{k=1}^K W_{ik} \hat{Y}_{ik} \\
    \nonumber \textit{s.t.}& \quad W_i \ge W_{i1} \ge W_{i2} \ge \cdots \ge W_{iK} \\
    \nonumber    &\quad W_i + \Sigma_{k=1}^K W_{ik} = 1,
\end{align}
where the $i$-th token is the \textit{anchor} token and $K$ \textit{matched} tokens are selected according to the KNN strategy.
We sort the $K$ \textit{matched} tokens in ascending order according to the distance to the \textit{anchor} token in the old feature space.
$\hat{Y}_i$, $\hat{Y}_{ik}$ are the prediction scores of the $i$-th token and its $k$-th \textit{matched} token, and $W_i$, $W_{ik}$ are the weight for $\hat{Y}_i$, $\hat{Y}_{ik}$, respectively.
The weight constraints ensure that the token closer to the \textit{colliding} feature has a more significant effect.

Until now, we calculate the effects in $D^E$ and $D^O$ and ignore the \textit{Undefined-Other}-Class.
Following \citet{monaikul2021continual}, we apply the standard knowledge distillation to allow new models to learn from old models.
To this end, we just need to re-write $\textit{Effect}_O$ in Eq.(\ref{eq:total_objective_1}) as follow:
\begin{align}
    \textit{Effect}_O &=  - \Sigma_j \textit{KL}(\overline{Y}_j,\widetilde{Y}_j) - \Sigma_n \textit{KL}(\hat{Y}_n,\widetilde{Y}_n) \label{eq:effect_O} \\
    \nonumber \textit{s.t.}& \quad D_j \in D^O, \quad D_n \in D^{UO},
\end{align}
where $D^{UO}$ is the data belong to the \textit{Undefined-Other}-Class.
The Eq.(\ref{eq:effect_O}) can be seen as calculating samples from $D^O$ and $D^{UO}$ in the same way, except that samples from $D^{UO}$ have no \textit{matched} tokens. 
We summarize the proposed causal framework in Figure \ref{fig5}.

\subsection{Mitigating Label Noise in $\textit{Effect}_O$\label{subsec:4.3}}
In our method, we use the old model to predict the labels of \textit{Defined-Other}-Class tokens.
However, it inevitably generates label noise when calculating $\textit{Effect}_O$.
To address this problem, we adopt curriculum learning to mitigate the label noise in the proposed method.

Curriculum learning has been widely used for handling label noises \citep{guo2018curriculumnet} in computer vision.
\citet{arazo2019unsupervised} empirically find that networks tend to fit correct samples before noisy samples.
Motivated by this,  we introduce a confidence threshold $\delta$ ($\delta \in [0,1]$) to encourage the model to learn first from clean \textit{Other}-class samples and then noisier ones.
Specifically, when calculating $\textit{Effect}_O$, we only select \textit{Defined-Other}-Class tokens whose predictive confidences are larger than $\delta$ for distilling colliding effects while others are for knowledge distillation.
The value of $\delta$ changes along with the training process and the value of $\delta$ in the $i$-th epoch is calculated as follow:
\begin{equation}
    \delta_i = 
    \begin{cases}
    \delta_1 + \frac{i-1}{m-1}(\delta_m-\delta_1), \quad 1 \le i \le m \\
    \delta_m ,\quad i > m, 
    \end{cases}
\end{equation}
where $m$, $\delta_1$ and $\delta_m$ are the predefined hyper-parameters and $\delta_m$ should be smaller than $\delta_1$.

\subsection{Balancing $\textit{Effect}_E$ and $\textit{Effect}_O$ \label{subsec:4.4}}
Figure \ref{fig5} shows that the total causal effect $\textit{Effect}$ consists of $\textit{Effect}_E$ and $\textit{Effect}_O$, where $\textit{Effect}_E$ is for learning new entities while $\textit{Effect}_O$ is for reviewing old knowledge.
With the learning process of CL, the need to preserve old knowledge varies \citep{hou2019learning}.
For example, more efforts should be made to preserve old knowledge when there are 15 old classes and 1 new class \textit{v.s.} 5 old classes and 1 new class.
In response to this, we introduce a self-adaptive weight for balancing $\textit{Effect}_E$ and $\textit{Effect}_O$:
\begin{equation}
    \lambda = \lambda_{base}\sqrt{C_O/C_N}
\end{equation}
where $\lambda_{base}$ is the initial weight and $C_O$,$C_N$ are the numbers of old and new entity types respectively.
In this way, the causal effects from new entity types and \textit{Other}-class are dynamically balanced when the ratio of old classes to new classes changes.
Finally, the objective of the proposed method is given as follow:
\begin{equation}
    \max \textit{Effect} = \textit{Effect}_E + \lambda * \textit{Effect}_O
\end{equation}

\begin{table*}[!t]
\centering
\caption{Comparisons with state-of-the-art methods on I2B2 and OntoNotes5. The average results as well as standard derivations are provided. \textit{Mi-F1}: micro-F1; \textit{Ma-F1}: macro-F1; \textit{Forget}: Forgetting; : higher is better; : lower is better. The best F1 results are bold.}
\resizebox{0.9\linewidth}{!}{%
\begin{tabular}{@{}cccccccccc@{}}
\toprule
 &  & \multicolumn{2}{c}{FG-1-PG-1} & \multicolumn{2}{c}{FG-2-PG-2} & \multicolumn{2}{c}{FG-8-PG-1} & \multicolumn{2}{c}{FG-8-PG-2} \\ \cmidrule(l){3-10} 
\multirow{-2}{*}{Dataset} & \multirow{-2}{*}{Method} & Mi-F1  & Ma-F1  & Mi-F1  & Ma-F1  & Mi-F1  & Ma-F1  & Mi-F1  & Ma-F1  \\ \midrule
 &  & \cellcolor[HTML]{EFEFEF}17.43 & \cellcolor[HTML]{EFEFEF}13.81 &  \cellcolor[HTML]{EFEFEF}28.57 & \cellcolor[HTML]{EFEFEF}21.43 &  \cellcolor[HTML]{EFEFEF}20.83 & \cellcolor[HTML]{EFEFEF}18.11 &  \cellcolor[HTML]{EFEFEF}23.60 & \cellcolor[HTML]{EFEFEF}23.54\\
 & \multirow{-2}{*}{Finetune Only} & \cellcolor[HTML]{EFEFEF}±0.54 & \cellcolor[HTML]{EFEFEF}±1.14 & \cellcolor[HTML]{EFEFEF}±0.26 & \cellcolor[HTML]{EFEFEF}±0.41 & \cellcolor[HTML]{EFEFEF}±1.78 & \cellcolor[HTML]{EFEFEF}±1.66 & \cellcolor[HTML]{EFEFEF}±0.15 & \cellcolor[HTML]{EFEFEF}±0.38\\
 &  & 12.31 & 17.14 & 34.67 & 24.62& 39.26 & 27.23& 36.22 & 26.08\\
 & \multirow{-2}{*}{PODNet} & ±0.35 & ±1.03 & ±2.65 & ±1.76 & ±1.38 & ±0.93 & ±12.9 & ±7.42 \\
 &  & \cellcolor[HTML]{EFEFEF}43.86 & \cellcolor[HTML]{EFEFEF}31.31 &  \cellcolor[HTML]{EFEFEF}64.32 & \cellcolor[HTML]{EFEFEF}43.53 &  \cellcolor[HTML]{EFEFEF}57.86 & \cellcolor[HTML]{EFEFEF}33.04 &  \cellcolor[HTML]{EFEFEF}68.54 & \cellcolor[HTML]{EFEFEF}46.94\\
 & \multirow{-2}{*}{LUCIR} & \cellcolor[HTML]{EFEFEF}±2.43 & \cellcolor[HTML]{EFEFEF}±1.62 &  \cellcolor[HTML]{EFEFEF}±0.76 & \cellcolor[HTML]{EFEFEF}±0.59 &  \cellcolor[HTML]{EFEFEF}±0.87 & \cellcolor[HTML]{EFEFEF}±0.39 & \cellcolor[HTML]{EFEFEF}±0.27 & \cellcolor[HTML]{EFEFEF}±0.63\\
 &  & 31.98 & 14.76 & 55.44 & 33.38 & 49.51 & 23.77 & 48.94 & 29.00\\
 & \multirow{-2}{*}{ST} & ±2.12 & ±1.31 & ±4.78 & ±3.13 & ±1.35 & ±1.01 & ±6.78 & ±3.04 \\
 &  & \cellcolor[HTML]{EFEFEF}42.85 & \cellcolor[HTML]{EFEFEF}24.05 & \cellcolor[HTML]{EFEFEF}57.01 & \cellcolor[HTML]{EFEFEF}35.29 & \cellcolor[HTML]{EFEFEF}43.95 & \cellcolor[HTML]{EFEFEF}23.12 & \cellcolor[HTML]{EFEFEF}52.25 & \cellcolor[HTML]{EFEFEF}30.93 \\
 & \multirow{-2}{*}{ExtendNER} & \cellcolor[HTML]{EFEFEF}±2.86 & \cellcolor[HTML]{EFEFEF}±1.35 & \cellcolor[HTML]{EFEFEF}±4.14 & \cellcolor[HTML]{EFEFEF}±3.38 & \cellcolor[HTML]{EFEFEF}±2.01 & \cellcolor[HTML]{EFEFEF}±1.79 & \cellcolor[HTML]{EFEFEF}±5.36 & \cellcolor[HTML]{EFEFEF}±2.77\\
 &  & \textbf{62.73} & \textbf{36.26} & \textbf{71.98} & \textbf{49.09} & \textbf{59.79} & \textbf{37.3} & \textbf{69.07} & \textbf{51.09} \\
\multirow{-12}{*}{I2B2} & \multirow{-2}{*}{CFNER(Ours)} & \textbf{±3.62} & \textbf{±2.24} & \textbf{±0.50} & \textbf{±1.38} & \textbf{±1.70} & \textbf{±1.15} & \textbf{±0.89} & \textbf{±1.05} \\ \midrule
 &  & \cellcolor[HTML]{EFEFEF}15.27 & \cellcolor[HTML]{EFEFEF}10.85 & \cellcolor[HTML]{EFEFEF}25.85 & \cellcolor[HTML]{EFEFEF}20.55 & \cellcolor[HTML]{EFEFEF}17.63 & \cellcolor[HTML]{EFEFEF}12.23 & \cellcolor[HTML]{EFEFEF}29.81 & \cellcolor[HTML]{EFEFEF}20.05\\
 & \multirow{-2}{*}{Finetune Only} & \cellcolor[HTML]{EFEFEF}±0.26 & \cellcolor[HTML]{EFEFEF}±1.11 & \cellcolor[HTML]{EFEFEF}±0.11 & \cellcolor[HTML]{EFEFEF}±0.24 & \cellcolor[HTML]{EFEFEF}±0.57 & \cellcolor[HTML]{EFEFEF}±1.08 & \cellcolor[HTML]{EFEFEF}±0.12 & \cellcolor[HTML]{EFEFEF}±0.16 \\
 &  & 9.06 & 8.36 & 34.67 & 24.62 & 29.00 & 20.54 & 37.38 & 25.85 \\
 & \multirow{-2}{*}{PODNet} & ±0.56 & ±0.57 & ±1.08 & ±0.85 & ±0.86 & ±0.91 & ±0.26 & ±0.29 \\
 &  & \cellcolor[HTML]{EFEFEF}28.18 & \cellcolor[HTML]{EFEFEF}21.11 & \cellcolor[HTML]{EFEFEF}64.32 & \cellcolor[HTML]{EFEFEF}43.53 & \cellcolor[HTML]{EFEFEF}66.46 & \cellcolor[HTML]{EFEFEF}46.29 & \cellcolor[HTML]{EFEFEF}76.17 & \cellcolor[HTML]{EFEFEF}55.58 \\
 & \multirow{-2}{*}{LUCIR} & \cellcolor[HTML]{EFEFEF}±1.15 & \cellcolor[HTML]{EFEFEF}±0.84 & \cellcolor[HTML]{EFEFEF}±1.79 & \cellcolor[HTML]{EFEFEF}±1.11 & \cellcolor[HTML]{EFEFEF}±0.46 & \cellcolor[HTML]{EFEFEF}±0.38 & \cellcolor[HTML]{EFEFEF}±0.09 & \cellcolor[HTML]{EFEFEF}±0.55\\
 &  & 50.71 & 33.24 & 68.93 & 50.63 & 73.59 & 49.41 & 77.07 & 53.32 \\
 & \multirow{-2}{*}{ST} & ±0.79 & ±1.06 & ±1.67 & ±1.66 & ±0.66 & ±0.77 & ±0.62 & ±0.63 \\
 &  & \cellcolor[HTML]{EFEFEF}50.53 & \cellcolor[HTML]{EFEFEF}32.84 & \cellcolor[HTML]{EFEFEF}67.61 & \cellcolor[HTML]{EFEFEF}49.26 & \cellcolor[HTML]{EFEFEF}73.12 & \cellcolor[HTML]{EFEFEF}49.55 & \cellcolor[HTML]{EFEFEF}76.85 & \cellcolor[HTML]{EFEFEF}54.37 \\
 & \multirow{-2}{*}{ExtendNER} & \cellcolor[HTML]{EFEFEF}±0.86 & \cellcolor[HTML]{EFEFEF}±0.84 & \cellcolor[HTML]{EFEFEF}±1.53 & \cellcolor[HTML]{EFEFEF}±1.49 & \cellcolor[HTML]{EFEFEF}±0.93 & \cellcolor[HTML]{EFEFEF}±0.90 & \cellcolor[HTML]{EFEFEF}±0.77 & \cellcolor[HTML]{EFEFEF}±0.57\\
 &  & \textbf{58.94} & \textbf{42.22} & \textbf{72.59} & \textbf{55.96} & \textbf{78.92} & \textbf{57.51} & \textbf{80.68} & \textbf{60.52} \\
\multirow{-12}{*}{OntoNotes5} & \multirow{-2}{*}{CFNER(Ours)} & \textbf{±0.57} & \textbf{±1.10} & \textbf{±0.48} & \textbf{±0.69} & \textbf{±0.58} & \textbf{±1.32} & \textbf{±0.25} & \textbf{±0.84} \\ \bottomrule
\end{tabular}
}
\label{tab:main_result_1_full}
\end{table*}

\begin{figure*}[!t]
    \centering
    \subfloat[I2B2 (FG-1-PG-1)]{
        \includegraphics[width=0.26\linewidth]{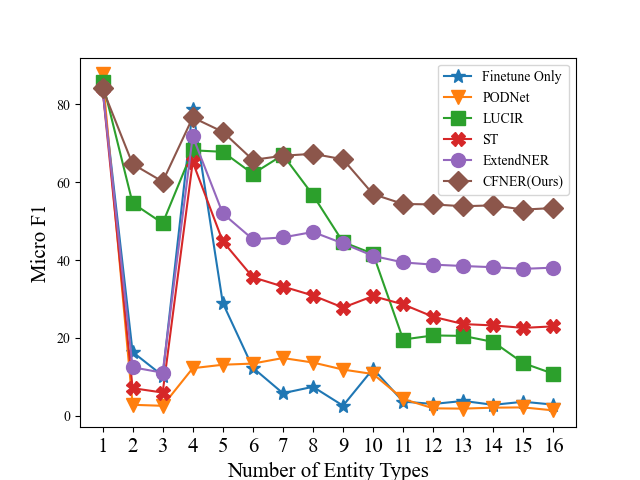}
    }\hspace{-0.6cm}\vspace{-0.4cm}
    \subfloat[I2B2 (FG-2-PG-2)]{
        \includegraphics[width=0.26\linewidth]{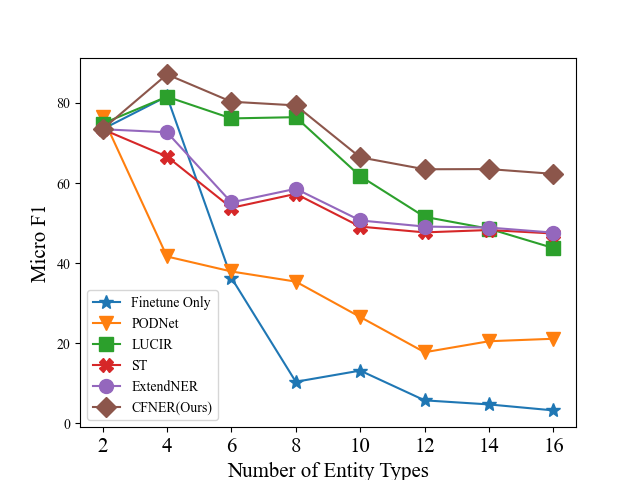}
    }\hspace{-0.6cm}
    \subfloat[I2B2 (FG-8-PG-1)]{
        \includegraphics[width=0.26\linewidth]{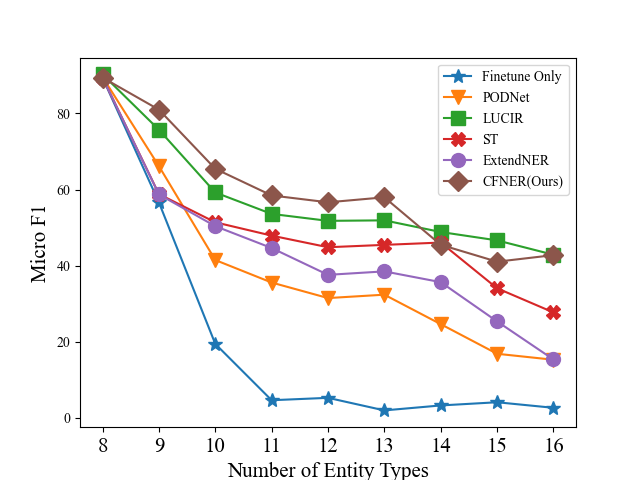}
    }\hspace{-0.6cm}
    \subfloat[I2B2 (FG-8-PG-2)]{
        \includegraphics[width=0.26\linewidth]{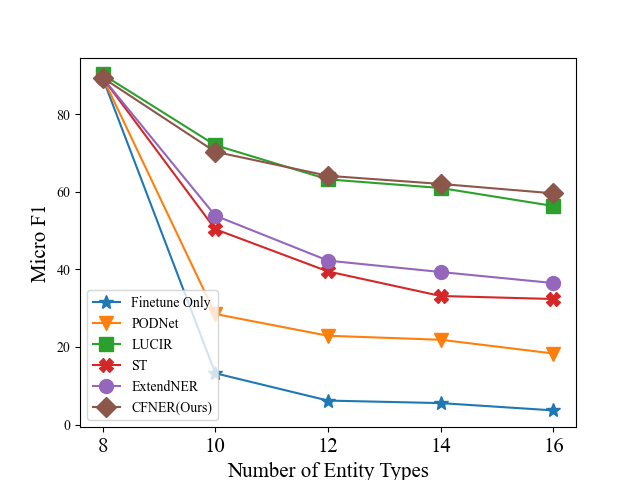}
    }\hspace{-0.6cm}
    \subfloat[OntoNotes5 (FG-1-PG-1)]{
        \includegraphics[width=0.26\linewidth]{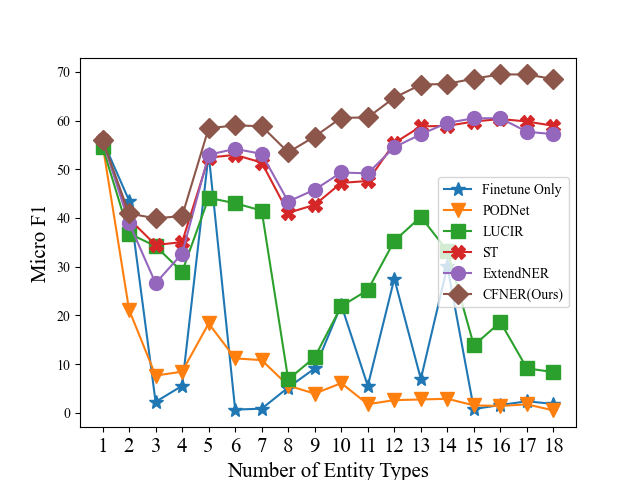}
    }\hspace{-0.6cm}
    \subfloat[OntoNotes5 (FG-2-PG-2)]{
        \includegraphics[width=0.26\linewidth]{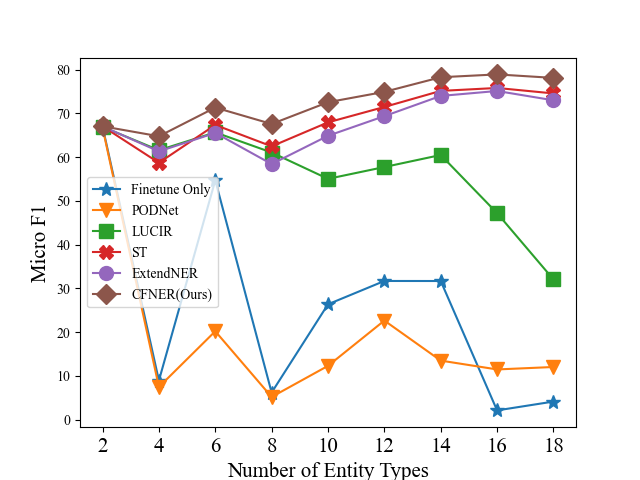}
    }\hspace{-0.6cm}
    \subfloat[OntoNotes5 (FG-8-PG-1)]{
        \includegraphics[width=0.26\linewidth]{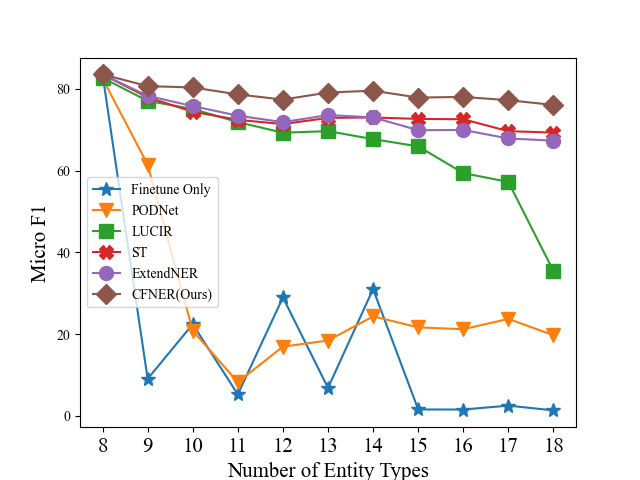}
    }\hspace{-0.6cm}
    \subfloat[OntoNotes5 (FG-8-PG-2)]{
        \includegraphics[width=0.26\linewidth]{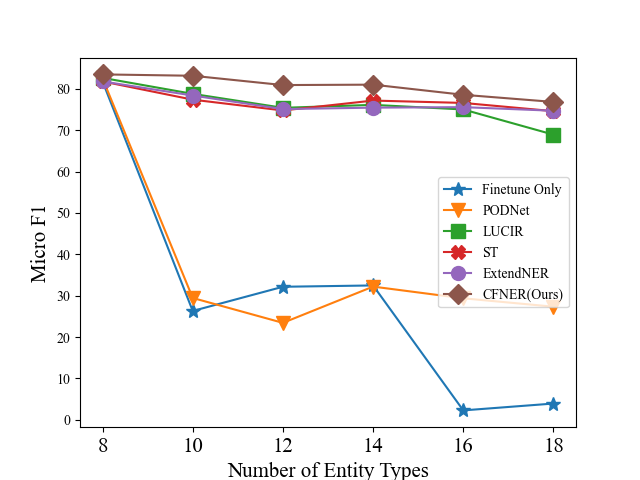}
    }
    \caption{Comparison of the step-wise micro-f1 score on I2B2 (16 entity types), OntoNotes5 (18 entity types). }
    \label{fig:main_result_step_1}
\end{figure*}

\section{Experiments}
\subsection{Settings}
\noindent\textbf{Datasets.}\quad 
We conduct experiments on three widely used datasets, \textit{i.e.}, OntoNotes5 \citep{hovy2006ontonotes}, i2b2 \citep{murphy2010serving} and CoNLL2003 \citep{sang2003introduction}.
To ensure that each entity type has enough samples for training, we filter out the entity types which contain less than 50 training samples.
We summarize the statistics of the datasets in Table \ref{tab:dataset_statistics} in Appendix \ref{appendix:additional_results}.

Following \citet{monaikul2021continual}, we split the training set into disjoint slices, and in each slice, we only retain the labels which belong to the entity types to learn while setting other labels to \textit{Other}-class.
Different from \citet{monaikul2021continual}, we adopt a greedy sampling strategy to partition the training set to better simulate the real-world scenario.
Specifically, the sampling algorithm encourages that the samples of each entity type are mainly distributed in the slice to learn.
We provide more explanations and the detailed algorithm in Appendix \ref{appendix:greedy_algorithm}. 

\noindent\textbf{Training.}\quad
We use bert-base-cased \citep{devlin-etal-2019-bert} as the backbone model and a fully-connected layer for classification.
Following previous work in CL \citep{hu2021distilling}, we predefine a fixed order of classes (alphabetically in this study) and train models with the corresponding data slice sequentially.
Specifically, $FG$ entity types are used to train a model as the initial model, and every $PG$ entity types are used for training in each CL step (denoted as $FG$-$a$-$PG$-$b$).
For evaluation, we only retain the new entity types' labels while setting other labels to \textit{Other}-class in the validation set. 
In each CL step, we select the model with the best validation performance for testing and the next step's learning. 
For testing, we retain the labels of all recognized entity types while setting others to \textit{Other}-class in the test set.

\noindent\textbf{Metrics.}\quad
Considering the class imbalance problem in NER, we adopt Micro F1 and Macro F1 for measuring the model performance.
We report the average result on all CL steps (including the first step) as the final result.

\noindent\textbf{Baselines.}\quad
We consider four baselines: 
ExtendNER \citep{monaikul2021continual}, Self-Training (ST) \citep{rosenberg2005semi,de2019continual}, LUCIR \citep{hou2019learning} and PODNet \citep{douillard2020podnet}.
ExtendNER is the previous state-of-the-art method in CL-NER.
LUCIR and PODNet are state-of-the-art CL methods in computer vision.
Detailed descriptions of the baselines and their training settings are demonstrated in Appendix \ref{baselines_settings}.

\noindent\textbf{Hyper-Parameters.}\quad
We set the number of matched tokens $K=3$, the weights $W_i=1/2$ and $W_{ik}=\frac{1}{2K}$.
For parameters in the curriculum learning strategy, we set $\delta_1=1, \delta_m=0$ and $m=10$.
We set the initial value of balancing weight $\lambda_{base}=2$.
More training details are shown in Appendix \ref{appendix:training_details}.

\subsection{Results and Analysis}
\noindent\textbf{Comparisons with State-Of-The-Art.}\quad
We consider two scenarios for each dataset: (1) training the first model the same as the following CL steps; (2) training the first model with half of all entity types.
The former scenario is more challenging, whereas the latter is closer to the real-world scenario since it allows models to learn enough knowledge before incremental learning.
Apart from that, we consider fine-tuning without any anti-forgetting techniques (Finetune Only) as a lower bound for comparison.

The results on I2B2 and OntoNotes5 are summarized in Table \ref{tab:main_result_1_full} and Figure \ref{fig:main_result_step_1}.
Due to the space limitation, we provide the results on CoNLL2003 in Table \ref{tab:main_result_2_full} and Figure \ref{fig:main_result_step_2} in Appendix \ref{appendix:additional_results}.
In most cases, our method achieves the best performance.
Especially, our method outperforms the previous state-of-the-art method in CL-NER (\textit{i,e.,} ExtendNER) by a large margin.
Besides, we visualize the features of our method and ExtendNER for comparison in Appendix \ref{appendix:tsne_visualization}.
The performances of PODNet and LUCIR are much worse than our methods when more CL steps are performed. 
The reason could be that neither of them differentiates \textit{Other}-class from entity types, and the old knowledge in \textit{Other}-class is not preserved.
Our method encourages the model to review old knowledge from both new entity types and \textit{Other}-class in the form of distilling causal effects.

\begin{table}[htbp]
  \tiny
  \centering
  \caption{The ablation study of our method on three datasets in the setting FG-1-PG-1. \textit{AW}: adaptive weight; \textit{CuL}: curriculum learning strategy; \textit{Mi-F1}: micro-F1; \textit{Ma-F1}: macro-F1.}
  \resizebox{\linewidth}{!}{
    \begin{tabular}{lcccccc}
    \toprule
    \multirow{2}[4]{*}{Methods} & \multicolumn{2}{c}{I2B2} & \multicolumn{2}{c}{OntoNotes5} & \multicolumn{2}{c}{CoNLL2003} \\
\cmidrule{2-7}          & Mi-F1 & Ma-F1 & Mi-F1 & Ma-F1 & Mi-F1 & Ma-F1 \\
    \midrule
    CFNER(Ours)  & \textbf{62.73}  & \textbf{36.26}  & \textbf{58.94}  & \textbf{42.22}  & \textbf{80.91}  & \textbf{79.11}  \\
    w/o AW & 61.65  & 35.86  & 57.63  & 40.28  & 80.75  & 78.43  \\
    w/o CuL & 61.21  & 34.79  & 57.95  & 39.54  & 80.32  & 78.71  \\
    w/o AW \& CuL
      & 60.78  & 33.15  & 56.83  & 38.95  & 79.89  & 77.54  \\
    w/o $\textit{Effect}_O$ & 59.68  & 30.56  & 53.09  & 34.99  & 78.68  & 76.15  \\
    w/o $\textit{Effect}_E$ & 53.62  & 28.75  & 54.88  & 37.29  & 79.83  & 77.45  \\
    w/o $\textit{Effect}_O$ \& $\textit{Effect}_E$ & 42.85  & 24.05  & 50.53  & 32.84  & 76.36  & 73.04  \\
    \bottomrule
    \end{tabular}%
    }
  \label{tab:ablation_study}%
\end{table}%

\noindent\textbf{Ablation Study.}\quad
We ablate our method, and the results are summarized in Table \ref{tab:ablation_study}.
To validate the effectiveness of the proposed causal framework, we only remove the colliding effects in \textit{Other}-class and new entity types for the settings w/o $\textit{Effect}_O$ and w/o $\textit{Effect}_E$, respectively.
Specifically, in the w/o $\textit{Effect}_O$ setting, we apply knowledge distillation for all \textit{Other}-class samples, while in w/o $\textit{Effect}_E$ setting, we calculate the cross-entropy loss for classification.
Note that our model is the same as ExtendNER when no causal effects are used (\textit{i.e.,} w/o $\textit{Effect}_O$ \& $\textit{Effect}_E$).
The results show that both $\textit{Effect}_O$ and $\textit{Effect}_E$ play essential roles in our framework.
Furthermore, the adaptive weight and the curriculum-learning strategy help model better learn causal effects in new data.

\begin{table}[htbp]
  \centering
  \caption{Hyper-parameter analysis on I2B2 (FG-8-PG-2). \textit{Mi-F1}: micro-F1; \textit{Ma-F1}: macro-F1.}
    \resizebox{\linewidth}{!}{
    \begin{tabular}{c|cc||c|cc||c|cc}
    \toprule
    $K$   & Mi-F1 & Ma-F1 & $\delta_1$ & Mi-F1 & Ma-F1 & $\lambda_{base}$ & Mi-F1 & Ma-F1 \\
    \midrule
    1     & 65.48 & 46.82 & 0     & 65.25 & 46.33 & 0.5   & 69.27 & 48.71 \\
    2     & 67.12 & 49.37 & 0.5   & 66.09 & 48.51 & 1     & 69.46 & 52.45 \\
    3     & 69.07 & 51.09 & 0.9   & 68.23 & 49.1  & 2     & 69.07 & 51.09 \\
    5     & 70.25 & 52.51 & 0.95  & 68.64 & 49.66 & 5     & 64.4  & 46.65 \\
    10    & 70.69 & 52.26 & 1     & 69.07 & 51.09 & 10    & 54.12 & 40.66 \\
    \bottomrule
    \end{tabular}%
    }
  \label{tab:hyperparameters_analysis}%
\end{table}%

\noindent\textbf{Hyper-Parameter Analysis.}\quad
We provide hyper-parameter analysis on I2B2 with the setting FG-8-PG-2.
We consider three hyper-parameters: the number of matched tokens $K$, the initial value of balancing weight $\lambda_{base}$ and the initial value of confidence threshold $\delta_1$.
The results in Table \ref{tab:hyperparameters_analysis} shows that a larger $K$ is beneficial.
However, as $K$ becomes larger, the run time increases correspondingly.
The reason is that more forward passes are required during training.
Therefore, We select $K=3$ by default to balance effectiveness and efficiency.
Results also show that $\delta_1=1$ reaches the best result, which indicates that it is more effective to learn $\textit{Effect}_E$ first and then gradually introduce $\textit{Effect}_O$ during training.
Otherwise, the old model's wrong predictions will significantly affect the model's performance.
Additionally, we find that the performance drops substantially when $\lambda_{base}$ is too large. 
Note that we did not carefully search for the best hyper-parameters, and the default ones are used throughout the experiments.
Therefore, elaborately adjusting the hyper-parameters may lead to superior performances on specific datasets and scenarios.

\begin{table}[htbp]
  \centering
  \caption{Combining our methods with other baselines on three datasets in the setting FG-1-PG-1. \textit{Mi-F1}: micro-F1; \textit{Ma-F1}: macro-F1. \textit{CF} represents applying causal effects.}
  \resizebox{\linewidth}{!}{
        \begin{tabular}{ccccccc}
    \toprule
    \multirow{2}[4]{*}{Methods} & \multicolumn{2}{c}{I2B2} & \multicolumn{2}{c}{OntoNotes5} & \multicolumn{2}{c}{CoNLL2003} \\
\cmidrule{2-7}          & Mi-F1 & Ma-F1 & Mi-F1 & Ma-F1 & Mi-F1 & Ma-F1 \\
    \midrule
    CFNER(Ours)  & 62.73  & 36.26  & 58.94  & 42.22  & 80.91 & 79.11 \\
    \midrule
    LUCIR & 43.86  & 31.31  & 28.18  & 21.11  & 74.15 & 70.48 \\
    LUCIR+CF & 66.27  & 38.52  & 62.03  & 44.34  & 81.28  & 79.56  \\
    \midrule
    ST    & 31.98  & 14.76  & 50.71  & 33.24  & 76.17  & 72.88  \\
    ST+CF & 61.41  & 33.43  & 57.06  & 41.28  & 80.59  & 79.11  \\
    \bottomrule
    \end{tabular}%
    }
  \label{tab:combine_other_baselines}%
\end{table}%

\noindent\textbf{Combining with Other Baselines.}\quad
Furthermore, the proposed causal framework can be implemented as a plug-and-play module (denoted as \textit{CF}).
As shown in Figure \ref{fig5}, $\textit{Effect}_E$ is based on the cross-entropy loss for classifying new entities, while $\textit{Effect}_O$ is based on the KL divergence loss for the prediction-level distillation.
We use LUCIR and ST as the baselines for a demonstration.
To combine LUCIR with our method, we substitute the classification loss in LUCIR with $\textit{Effect}_E$ and substitute the feature-level distillation with $\textit{Effect}_O$.
When combining with ST, we only replace the soft label with the hard label when calculating $\textit{Effect}_O$.
The results in Table \ref{tab:combine_other_baselines} indicate that our method improves LUCIR and ST substantially.
It is worth noticing that \textit{LUCIR+CF} outperforms our method consistently, indicating our method has the potential to combine with other CL methods in computer vision and reach a superior performance in CL-NER. 
In addition, \textit{CFNER} outperforms \textit{ST+CF} due to the fact that soft labels convey more knowledge than hard labels.

\section{Conclusion}
How can we learn from \textit{Other}-class for anti-forgetting in CL-NER?
This question is answered by the proposed causal framework in this study.
Although \textit{Other}-class is ``useless'' in traditional NER, we point out that in CL-NER, \textit{Other}-class samples are naturally reserved samples from old classes, and we can distillate the causal effects from them to preserve old knowledge.
Moreover, we apply curriculum learning to alleviate the noisy label problem to enhance the distillation of $\textit{Effect}_O$.
We further introduce a self-adaptive weight for dynamically balancing causal effects between new entity types and \textit{Other}-class.
Experimental results show that the proposed causal framework not only outperforms state-of-the-art methods but also can be combined with other methods to further improve performances.

\section*{Ethics Statement}
For the consideration of ethical concerns, we would make description as follows:
(1) We conduct all experiments on existing datasets, which are derived from public scientific researches.
(2) We describe the statistics of the datasets and the hyper-parameter settings of our method. Our analysis is consistent with the experimental results.
(3) Our work does not involve sensitive data or sensitive tasks.
(4) We will open source our code for reproducibility.

\section*{Limitations}
Although the proposed method alleviates the catastrophic forgetting problem to some extent, its performances are still unsatisfactory when more CL steps are performed.
Additionally, calculating causal effects from \textit{Other}-class depends on the old model predictions, resulting in errors propagating to the following CL steps.
Moreover, the proposed method requires more computation and a longer training time since the predictions of \textit{matched} samples are calculated.


\bibliography{ref}
\bibliographystyle{acl_natbib}

\appendix

\section{Forgetting in CL-NER}
\label{appendix:forgetting}
To identify the cause of forgetting, we consider the differences in prediction $Y$ when the old data $S$ exists or not.
For each CL step, the effect of old data $S$ can be calculate as:
\begin{align}
    \textit{Effect}_{S} & = P(\hat{Y}=\hat{y}|\textit{do}(S=s)) \\
   \nonumber&- P(\hat{Y}=\hat{y}|\textit{do}(S=0)) \\
   &= P(\hat{Y}=\hat{y}|S=s) - P(\hat{Y}=\hat{y}|S=0) \\
   &= P(\hat{Y}=\hat{y}) - P(\hat{Y}=\hat{y}) \\
   &= 0,
   \label{eq:zero_effect}
\end{align}
where $\textit{do}(\cdot)$ is the \textit{causal intervention} \citep{pearl2014interpretation,pearl2009causality} representing that assigning a certain value to a variable without considering all parent nodes (causes).
In the first equation, $\textit{do}(S=0)$ represents null intervention, \textit{i.e.,} setting old data to null.
In the second equation, $P(\hat{Y}=\hat{y}|\textit{do}(S)=P(\hat{Y}=\hat{y}|S)$ due to the fact that $S$ has no parent nodes.
In the third equation, $P(\hat{Y}=\hat{y}|S)=P(\hat{Y}=\hat{y})$ since all causal paths from $S$ to $Y$ are blocked by the collider $X_0$.
From Eq.(\ref{eq:zero_effect}), we find that the missing old data effect causes the forgetting.
We neglect the effect of initial parameters adopted from the old model since it will be exponentially decayed towards 0 during learning \citep{kirkpatrick2017overcoming}.

\begin{figure*}[t]
    \centering
    \subfloat[Greedy Sampling]{
        \includegraphics[width=0.5\linewidth]{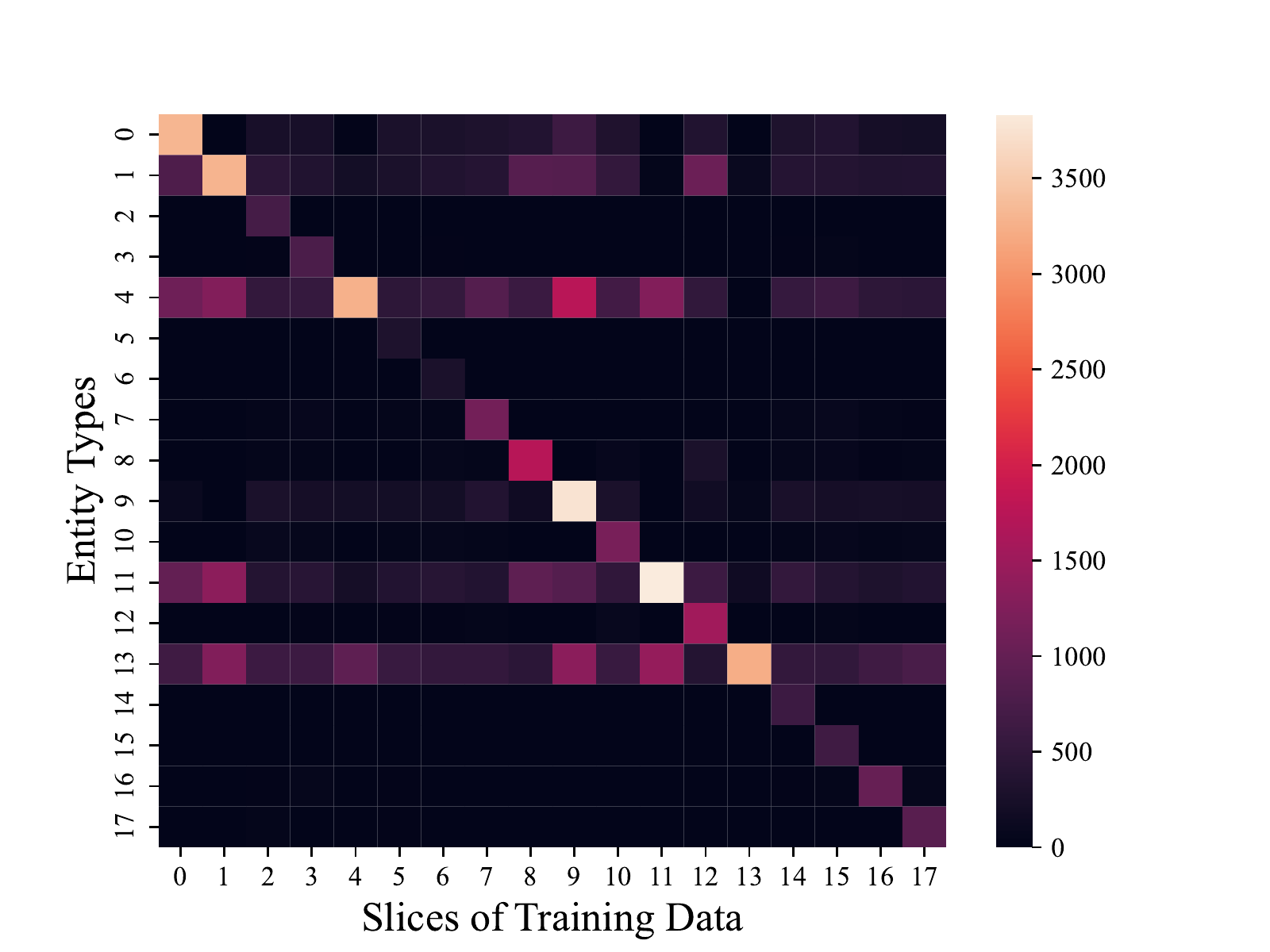}
    }
    \subfloat[Random Sampling]{
        \includegraphics[width=0.5\linewidth]{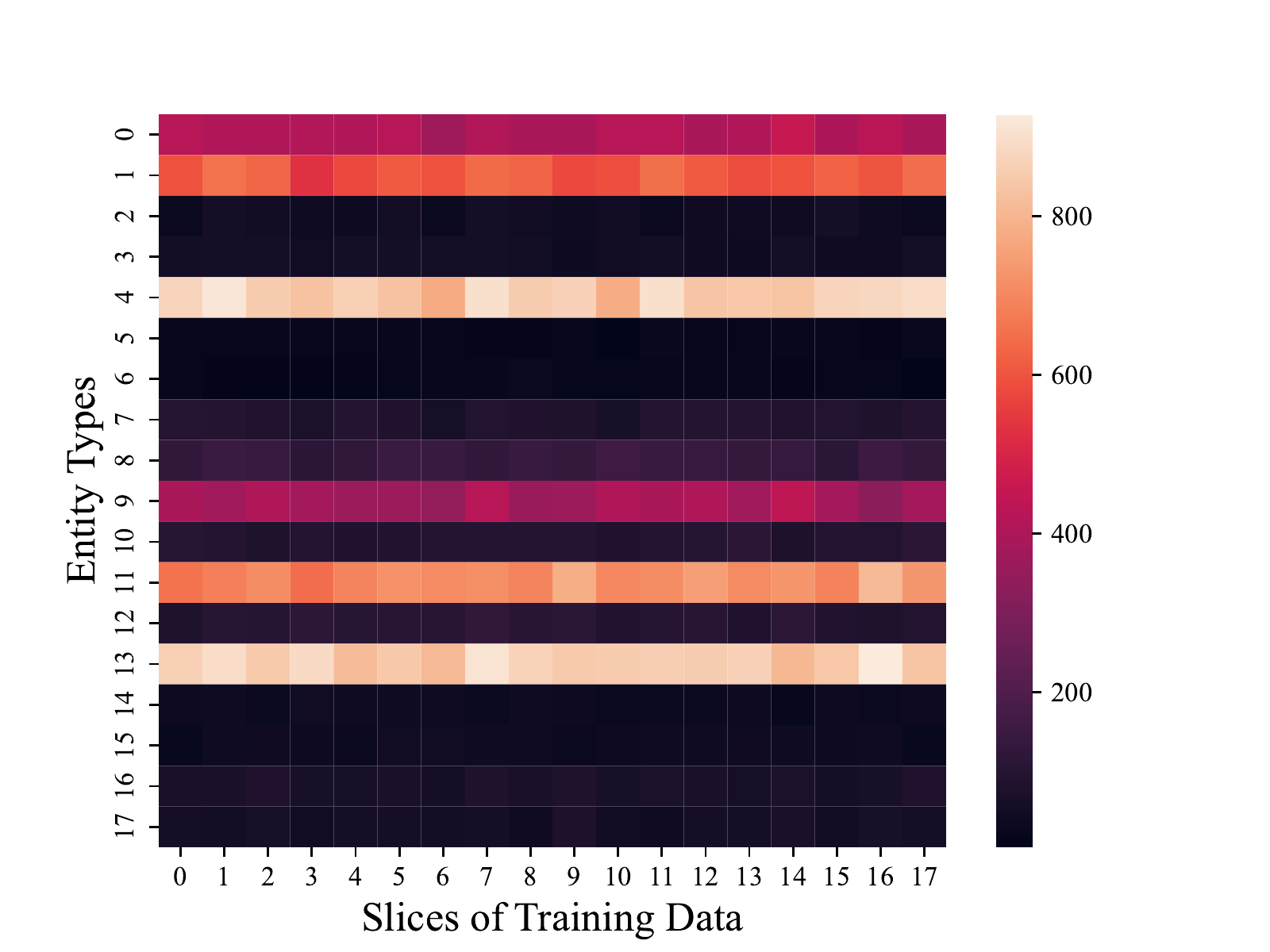}
    }
    \caption{Comparison of the greedy sampling and random sampling on OntoNotes5. Each slice contains one entity types to lean.}
    \label{fig:label_distri}
\end{figure*}

\SetKwInOut{KwResult}{Require}

\begin{algorithm}[!t] 
\caption{Greedy Sampling Algorithm for CL-NER} \label{greedy_sampling}
\KwIn{
$\mathcal{D}=\{(s_i,t_i)\}_{i=1}^{n}$: a training set contains $n$ sentences $s$ and their label sequences $t$; 
}
\KwResult{
$G$: the number of slices;
$\mathcal{E}$: the entity set;
$\mathcal{E}_j$: the entity set in the $j$-th slice;
}
\KwOut{
$G$ slices of datasets: $\mathcal{D}^{(1)},\mathcal{D}^{(2)},\cdots,\mathcal{D}^{(G)}$
}
Initialize $\mathcal{D}^{(1)},\cdots,\mathcal{D}^{(G)}=\{\},\cdots,\{\}$\;
Initialize $cnt_1,\cdots,cnt_G=0,\cdots,0$\;
\tcp{Calculate the number of sentences to allocate in each slice}
\For{$j$ in range(1,G+1)}
{
 $n_j = n*|\mathcal{E}_j|/|\mathcal{E}|$
}
Sort $\mathcal{E}$ in ascending order based on the frequency in $\mathcal{D}$\;
\For{($s_i,t_i$) in $\mathcal{D}$}
{
    \textit{is\_select} = False\;
    \tcp{Allocate low frequency entity types in priority}
    \For{$e$ in $\mathcal{E}$}
    {
        \If{not $e$ in $t_i$}
        {
            continue\;
        }
        $j$ is index of $\mathcal{E}_j$ s.t. $e$ belongs to $\mathcal{E}_j$\;
        \If{$cnt_j<n_j$}
        {
            $\mathcal{D}^{(j)}=\mathcal{D}^{(j)}\cup{(s_i,t_i)}$\;
            $cnt_j$ +=1\;
            \textit{is\_select} = True\;
            break\;
        }
    }
    \tcp{Otherwise, the sentence is randomly assigned to an incomplete slice}
    \If{not is\_select}
    {
        randomly choice $j$ s.t. $cnt_j<n_j$\;
        $\mathcal{D}^{(j)}=\mathcal{D}^{(j)}\cup{(s_i,t_i)}$\;
        $cnt_j$ +=1\;
    }
}
\Return $\mathcal{D}^{(1)},\mathcal{D}^{(2)},\cdots,\mathcal{D}^{(G)}$\;
\end{algorithm}

\section{Greedy Sampling Algorithm for CL-NER\label{appendix:greedy_algorithm}}

In real-world scenarios, the new data should focus on the new entity types, \textit{i.e., } most sentences contain the tokens from new entity types.
Suppose we randomly partition a dataset for CL-NER as in \citet{monaikul2021continual}. 
In that case, each slice contains a large number of sentences whose tokens all belong to \textit{Other}-class, resulting in that models tend to bias to \textit{Other}-class when inference.
A straightforward solution is to filter out all sentences with only \textit{Other}-class tokens.
However, it brings a new problem: the slices' sizes are imbalanced.

To address this problem, we propose a sampling algorithm for partitioning a dataset in CL-NER (Algorithm \ref{greedy_sampling}).
Simply put, we allocate the sentence containing low-frequency entity types to the corresponding slice in priority until the slice contains the required number of sentences.
If a sentence contains no entity types or the corresponding slices are full, we randomly allocate the sentence to an incomplete slice.
In this way, we partition the dataset into slices with balanced sizes, and each slice mainly contains the entity types to learn.

For comparing Algorithm \ref{greedy_sampling} and the random sampling as in \citet{monaikul2021continual}, we provide the label distributions in each slices of training data in Figure \ref{fig:label_distri}.
Figure \ref{fig:label_distri} shows that the greedy sampling generates more realistic datasets for CL-NER.
When we use the randomly partitioned dataset for training in the setting FG-1-PG-1, the micro-f1 score of our method is 16.12, 17.43, and 12.75 (\%) on OntoNotes5, I2B2, and CoNLL2003, respectively, indicating that the number of entities in each slice is inadequate for learning a NER model.
Therefore, the greedy sampling alleviates the need for data in CL-NER.

\section{Baselines Introductions and Settings\label{baselines_settings}}
The introductions about the baselines in experiments and their experimental settings are as follows. 
Note that we do not apply any reserved samples from old classes in LUCIR and PODNet for a fair comparison since our method requires no old data.
\begin{itemize}
    \item Self-Training (ST) \citep{rosenberg2005semi,de2019continual}: 
    ST first utilizes the old model to annotate the \textit{Other}-class tokens with old entity types. 
    Then, the new model is trained on new data with annotations of all entity types. 
Finally, the cross-entropy loss on all entity types is minimized.
    \item ExtendNER \citep{monaikul2021continual}: 
    ExtendNER has a similar idea to ST, except that the old model provides the soft label (\textit{i.e.,} probability distribution) of \textit{Other}-class tokens. 
    Specifically, the cross-entropy loss is computed for entity types' tokens, and KL divergence loss is computed for \textit{Other}-class tokens. 
    During training, the sum of cross-entropy loss and KL divergence loss is minimized.
    Following \citet{monaikul2021continual}, we set the temperature of the teacher (old model) and student model (new model) to 1 and 2, respectively.
    \item LUCIR \citep{hou2019learning}: 
    LUCIR develops a framework for incrementally learning a unified classifier for the continual image classification tasks. 
    The total loss consists of three terms: 
    (1) the cross-entropy loss on the new classes samples; 
    (2) the distillation loss on the features extracted by the old model and those by the new one;
    (3) the margin-ranking loss on the reserved samples for old classes. 
    In our experiments, we compute the cross-entropy loss for new entity types, the distillation loss for all entity types, and the margin-ranking loss for \textit{Other}-class samples instead of the reserved samples. 
    Following \citep{hou2019learning}, $\lambda_{base}$ (\textit{i.e.,} loss weight for the distillation loss) is set to 50, \textit{K} (\textit{i.e.,} the top-K new class embeddings are chosen for the margin-ranking loss) is set to 1 and \textit{m} (\textit{i.e.,} the threshold of margin ranking loss) is set to 0.5 for all the tasks.
    \item PODNet \citep{douillard2020podnet}: 
    PODNet has a similar idea to LUCIR to combat the catastrophic forgetting in continual learning for image classification.
    The total loss consists of the classification loss and distillation loss. 
    To compute the distillation loss, PodNet constrains the output of each intermediate convolutional layer while LUCIR only considers the final feature embeddings. 
    For classification, PODNet used NCA loss instead of the cross-entropy loss. 
    In our experiments, we constrain the output of each intermediate stage of BERT as PODNet constrains each stage of a ResNet. 
    Following \citep{douillard2020podnet}, we set the $\lambda_{c}$ (\textit{i.e.,} loss weight for the POD-spatial loss) to 3 and $\lambda_{f}$ (\textit{i.e.,} loss weight for the POD-flat loss) to 1. 
\end{itemize}

\begin{figure*}[!t]
    \centering
    \subfloat[ExtendNER]{
        \includegraphics[width=0.4\linewidth]{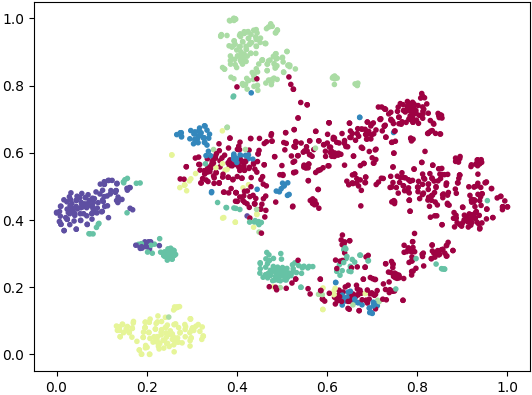}
    }
    \subfloat[CFNER(Ours)]{
        \includegraphics[width=0.4\linewidth]{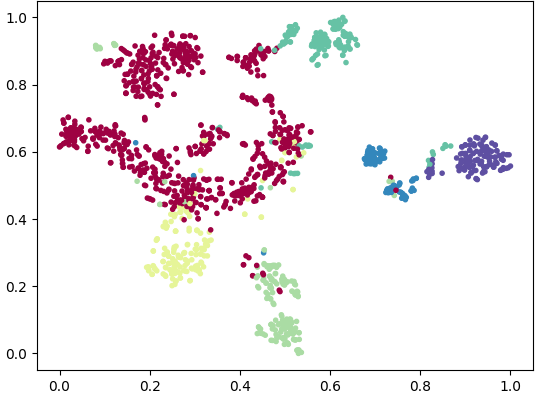}
    }
    \caption{The T-SNE visualization of the feature representations from BERT. The model is trained on OntoNotes5 in the setting FG-1-PG-1. We randomly select six classes for a demonstration.}
    \label{fig:tsen_visualization}
\end{figure*}

\begin{table}[!t]
  \centering
  \caption{The statistics and the entity sequence for each dataset}
  \resizebox{\linewidth}{!}{
    \begin{tabular}{cccccc}
    \toprule
          & \multicolumn{1}{l}{\# Class} & \# Sent & \# Entity & \multicolumn{2}{c}{Entity sequence (Alphabetical Order)} \\
    \midrule
    CoNLL2003 & 4     & 21k   & 35k   & \multicolumn{2}{l}{\begin{tabular}[1]{l}LOCATION, MISC, ORGANISATION, PERSON\end{tabular}} \\
    \midrule
    I2B2  & 16    & 141k  & 29k   & \multicolumn{2}{c}{\begin{tabular}[1]{l}AGE, CITY, COUNTRY, DATE, DOCTOR, HOSPITAL, \\  IDNUM, MEDICALRECORD, ORGANIZATION, \\PATIENT, PHONE, PROFESSION, STATE, STREET, \\USERNAME, ZIP\end{tabular}} \\
    \midrule
    OntoNotes5 & 18    & 77k   & 104k  & \multicolumn{2}{c}{\begin{tabular}[1]{l}CARDINAL, DATE, EVENT, FAC, GPE, LANGUAGE,\\ LAW, LOC, MONEY, NORP, ORDINAL, ORG,\\ PERCENT, PERSON, PRODUCT, QUANTITY, TIME,\\ WORK\_OF\_ART\end{tabular}} \\
    \bottomrule
    \end{tabular}%
    }
  \label{tab:dataset_statistics}%
\end{table}%

\section{Training Details \label{appendix:training_details}}

The models were implemented in Pytorch \cite{paszke2019pytorch} on top of the BERT Huggingface implementation \cite{wolf2019huggingface}.
We use the default hyper-parameters in BERT: hidden dimensions are 768, and the max sequence length is 512.
Following \citet{hu2021distilling}, we normalize the feature vector output by BERT and compute the cosine similarities between the feature vector and class representations for predictions.

We use the BIO tagging schema for all three datasets.
For CoNLL2003, we train the model for ten epochs in each CL step.
For OntoNotes5 and I2B2, we train the model for 20 epochs when PG=2, and 10 epochs when PG=1.
The batch size is set as 8, and the learning rate is set as 4e-4.
The experiments are run on GeForce RTX 2080 Ti GPU.
Each experiment is repeated 5 times.

\section{T-SNE Visualizations\label{appendix:tsne_visualization}}
To deepen the understanding of the forgetting in CL-NER, we visualize the feature representations from BERT in Figure \ref{fig:tsen_visualization}.
Results show that our method preserves more old knowledge and learns better feature representations than ExtendNER.

\section{Additional Experimental Results\label{appendix:additional_results}}

\begin{figure}[h]
    \centering
    \subfloat[CoNLL2003 (FG-1-PG-1)]{
        \includegraphics[width=\linewidth]{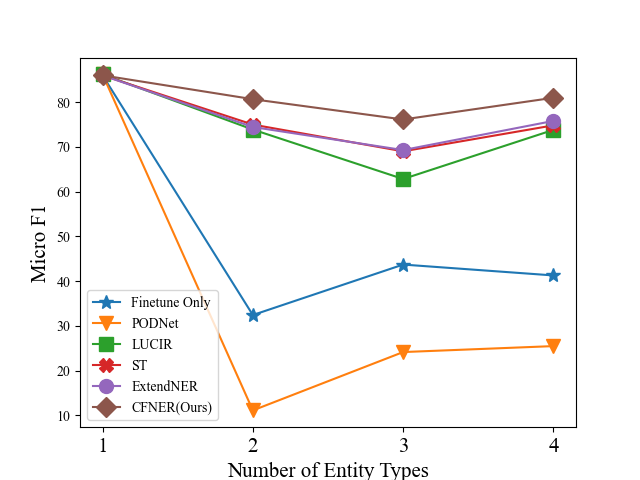}
    }
    \\
    \subfloat[CoNLL2003 (FG-2-PG-1)]{
        \includegraphics[width=\linewidth]{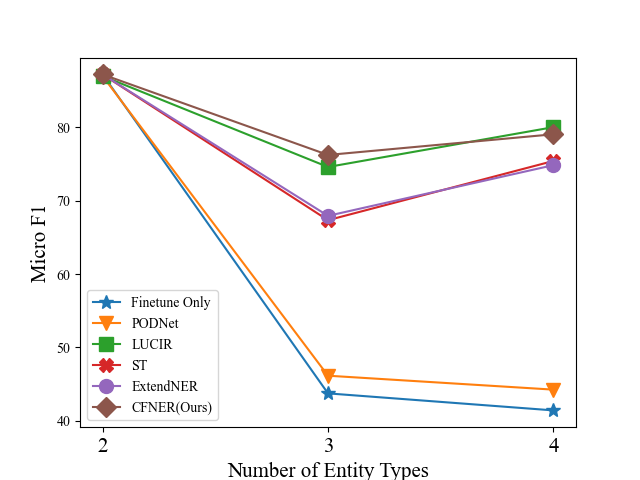}
    }
    \caption{Comparison of the step-wise micro-f1 score on CoNLL2003 (4 entity types).}
    \label{fig:main_result_step_2}
\end{figure}

\begin{table}[h]
\centering
\caption{Comparisons with state-of-the-art methods on CoNLL2003. The average results as well as standard derivations are provided. \textit{Mi-F1}: micro-F1; \textit{Ma-F1}: macro-F1. The best F1 results are bold.}
\resizebox{\linewidth}{!}{
\begin{tabular}{@{}cccccc@{}}
\toprule
 &  & \multicolumn{2}{c}{FG-1-PG-1} & \multicolumn{2}{c}{FG-2-PG-1} \\ \cmidrule(l){3-6} 
\multirow{-2}{*}{Dataset} & \multirow{-2}{*}{Method} & Mi-F1   & Ma-F1  & Mi-F1  & Ma-F1 \\ \midrule
 &  & \cellcolor[HTML]{EFEFEF}50.84 & \cellcolor[HTML]{EFEFEF}40.64 & \cellcolor[HTML]{EFEFEF}57.45 & \cellcolor[HTML]{EFEFEF}43.58\\
 & \multirow{-2}{*}{Finetune Only} & \cellcolor[HTML]{EFEFEF}±0.10 & \cellcolor[HTML]{EFEFEF}±0.16 & \cellcolor[HTML]{EFEFEF}±0.05 & \cellcolor[HTML]{EFEFEF}±0.18\\
 &  & 36.74 & 29.43 & 59.12 & 58.39 \\
 & \multirow{-2}{*}{PODNet} & ±0.52 & ±0.28 & ±0.54 & ±0.99 \\
 &  & \cellcolor[HTML]{EFEFEF}74.15 & \cellcolor[HTML]{EFEFEF}70.48 & \cellcolor[HTML]{EFEFEF}80.53 & \cellcolor[HTML]{EFEFEF}\textbf{77.33} \\
 & \multirow{-2}{*}{LUCIR} & \cellcolor[HTML]{EFEFEF}±0.43 & \cellcolor[HTML]{EFEFEF}±0.66 & \cellcolor[HTML]{EFEFEF}±0.31 & \cellcolor[HTML]{EFEFEF}\textbf{±0.31} \\
 &  & 76.17 & 72.88 & 76.65 & 66.72 \\
 & \multirow{-2}{*}{ST} & ±0.91 & ±1.12 & ±0.24 & ±0.11 \\
 &  & \cellcolor[HTML]{EFEFEF}76.36 & \cellcolor[HTML]{EFEFEF}73.04 & \cellcolor[HTML]{EFEFEF}76.66 & \cellcolor[HTML]{EFEFEF}66.36 \\
 & \multirow{-2}{*}{ExtendNER} & \cellcolor[HTML]{EFEFEF}±0.98 & \cellcolor[HTML]{EFEFEF}±1.8 & \cellcolor[HTML]{EFEFEF}±0.66 & \cellcolor[HTML]{EFEFEF}±0.64 \\
 &  & \textbf{80.91} & \textbf{79.11} & \textbf{80.83} & 75.20\\
\multirow{-12}{*}{CoNLL2003} & \multirow{-2}{*}{CFNER(Ours)} & \textbf{±0.29} & \textbf{±0.50} & \textbf{±0.36} & ±0.32 \\ \bottomrule
\end{tabular}
}
\label{tab:main_result_2_full}
\end{table}


\end{document}